\documentclass[10pt,twocolumn,letterpaper]{article}
\usepackage[normalem]{ulem}
\usepackage{cvpr}
\usepackage{times}
\usepackage{epsfig}
\usepackage{graphicx}
\usepackage{color,amsmath}
\usepackage{amssymb}
\usepackage{booktabs}
\usepackage[table]{xcolor}
\usepackage{subcaption}
\newcounter{savefootnote}

\usepackage{lipsum}

\newcommand{\ignore}[1]{}

\newcommand{\reals}{{\mathbb{R}}}

\newcommand{\targetsize}{{\mathcal{S}}}
\newcommand{\poolvar}{{\mathcal{V}}}
\newcommand{\weight}{{w^l}}
\newcommand{\vindex}{{i}}

\newcommand{\matvar}{{\hat{M}}}

\newcommand{\SMASH}{{\emph{SMH}}}

\usepackage[skip=6pt]{caption}

\usepackage[pagebackref=true,breaklinks=true,letterpaper=true,colorlinks,bookmarks=false]{hyperref}

\cvprfinalcopy %

\def\cvprPaperID{6403} %

\ifcvprfinal\fi

\makeatletter
\renewcommand*{\@fnsymbol}[1]{\ensuremath{\ifcase#1\or *\or \dagger\or \ddagger\or
    \mathsection\or *\or \|\or **\or \dagger\dagger
    \or \ddagger\ddagger \or 1 \else\@ctrerr\fi}}
\makeatother

\title{Structured Multi-Hashing for Model Compression}

\author{
    Elad Eban\thanks{The author contribute equally to this paper.\newline
    Elad and Yair contributed equally to the paper. They jointly proposed the idea of structured-multi-hashing. 
    Yair was the main contributor to the manuscript. 
    Elad wrote most of the code and ran EfficentNet experiments. 
    Hao contributed to coding and experiments. Yerlan ran CIFAR and ResNet experiments and simplified some aspects of the structured hashing. 
    Miguel advised Yerlan on issues about optimization and deep net compression. 
    Mark, and Andrew helped with MobileNet, and ResNet experiments.} \\
    Google Research. \\
    {\tt\small elade@google.com}
\and
    Yair Movshovitz-Attias\footnotemark[1]\\
    Google Research. \\
    {\tt\small yairmov@google.com}
\and
    Hao Wu\\
    Google. \\
    {\tt\small haou@google.com}
\and
    Mark Sandler\\
    Google Research. \\
    {\tt\small sandler@google.com}
\and
    Andrew Poon\\
    Google Research. \\
    {\tt\small ayp@google.com}
\and
    Yerlan Idelbayev\thanks{Worked performed while at Google Research.}\\
    University of California, Merced.\\
    {\tt\small yidelbayev@ucmerced.edu }
\and
    Miguel {\'A}. Carreira-Perpi{\~n}{\'a}n\\
    University of California, Merced.\\
    {\tt\small mcarreira-perpinan@ucmerced.edu}
}

\begin{document}

\setcounter{savefootnote}{\value{footnote}}%
\setcounter{footnote}{0}%

\maketitle

\begin{abstract}
    Despite the success of deep neural networks (DNNs), state-of-the-art models are too large to deploy on low-resource devices or common server configurations in which multiple models are held in memory. Model compression methods address this limitation by reducing the memory footprint, latency, or energy consumption of a model with minimal impact on accuracy. We focus on the task of reducing the number of learnable variables in the model.

    In this work we combine ideas from weight hashing and dimensionality reductions resulting in a simple and powerful structured multi-hashing method based on matrix products that allows direct control of model size of any deep network and is trained end-to-end.
    
    We demonstrate the strength of our approach by compressing models from the ResNet, EfficientNet, and MobileNet architecture families. Our method allows us to drastically decrease the number of variables while maintaining high accuracy.  For instance, by 
    applying our approach to EfficentNet-B4 (16M parameters) we reduce it to 
    to the size of B0 (5M parameters), while gaining over 3\% in accuracy over B0 baseline.
    
    On the commonly used benchmark CIFAR10 we reduce the ResNet32 model by 75\% with no loss in quality, and are able to do a 10x compression while still achieving above 90\% accuracy.
\end{abstract}

\section{Introduction}

The main factor driving the success of machine learning in recent years is the 
ability to build and train increasingly larger Deep Neural Networks (DNNs). 
This has been enabled by a combination of algorithmic advances such as ReLU activations~\cite{hahnloser2000digital, nair2010rectified}, Batch Normalization~\cite{ioffe2015batch}, and residual connections~\cite{he2016deep}; large training datasets~\cite{deng2009imagenet}; and faster, specialized, hardware~\cite{jouppi2017datacenter}.

Overwhelmingly, when given enough data, larger models show improvements in accuracy. However, this march upwards comes with a cost in terms of latency, energy, and memory consumption. For example, the popular Resnet-101 model~\cite{he2016deep} has 44 Million parameters and requires 150MB of storage;  AmoebaNet-A~\cite{real2019regularized} requires 469M parameters and 1800MB.
The size of DNNs limits their deployment in devices with low resources such as mobile phones and wearables. On server side, multi-tenancy -- the practice of serving multiple models from the same hardware accelerator -- is also affected by the model size. Furthermore, during inference, layers deeper in the network can be heavily affected by the cost of loading the weights.

From a scientific perspective, these models have many more parameters than the number of data points in the datasets they are trained on. This seems counter-intuitive as it seems to contradict learning-theory (e.g. VC dimension properties~\cite{vapnik2013nature}), but has been widely recognized as critical property of DNNs~\cite{Belkin2018ReconcilingMM, denton2014exploiting, ba2014deep}.
One wonders: Do the parameters of a network live in a lower dimensional space? Can we restrict the model class in a way that models in it can be represented efficiently (e.g. have low Kolmogorov Complexity) without sacrificing accuracy?
Can we find an intrinsic connection between the number of parameters of a model to its performance \cite{li2018measuring}? 

Note that low dimensionality assumptions are core in many CNN components. For example, convolutions are low dimensional linear maps and  separable convolutions (i.e.\ depthwise followed by $1\!\times\!1$ convolutions) are based on decomposition restrictions. However, making strong assumptions about individual elements in the model can be overly restrictive.

There is considerable interest in the machine learning community in making models cheaper: Reducing their size, either in number of parameters or as bytes on disk; lowering their latency; or reducing their memory and energy consumption during inference. Here we refer to these methods as \emph{model compression}.  

We can partition the model compression field into several types of techniques: 
architectural modifications, such as width multiplier, a move to separable convolutions, or filter number optimization~\cite{lecun1990optimal, gordon2018morphnet}; Neuron pruning, either during or after training; disk size compression~\cite{Oktay2019ModelCB}; weight quantization~\cite{Faraon_18a}; and hashing~\cite{weinberger2009feature, shi2009hash}. These approaches are in many ways complementary and have been used together~\cite{han2015deep}. In practice, hashing methods induce identity constraints between model weight that are mapped to the same variables. In addition, they lack memory locality which makes them slow and increases their RAM footprint. This has limited their adoption.

We present a new hashing approach for reducing the number of trainable variables in a model. We consider all weights in the DNN as if they are tiled into a single, large, matrix and represent it as a sum of products of multiple hashes, computed as matrix product. This defines a multi-hash from model weights into sets of trainable variables in which full collisions are exponentially rare, and are replaced by higher order correlations between weights. Using this representation, we then train the reduced model end to end.

We call this Structured Multi-Hashing (\SMASH).
\SMASH\ has a specific locality pattern which reduces cache misses and increases the efficiency of the compressed model. This representation is unique: it is not a linear subspace nor does it assume that any specific operation in the network is low rank.
Furthermore, by re-parameterizing hashing as a matrix product, the implementation becomes both simple and fast. It has little overhead in training or inference and results in much faster models compared to hashed models. We demonstrate the efficacy of \SMASH\ by applying it to state-of-the-art image classification models and drastically reducing their number of variables.

\section{Related Work}
Numerous efforts have been made on the topic of model compression, here we give a brief overview of different approaches.
\vspace{-1em}
    \paragraph{Hashing}
    The seminal work of Weinberger \etal~\cite{weinberger2009feature, shi2009hash} showed how useful hashing is in the context of linear classifiers. The work builds upon the kernel-trick and is designed to allow more efficient training and inference when the number of features and labels is huge.
    
    Chen \etal~\cite{chen2015compressing} extended this idea to the context of deep networks introducing HashNets. Each layer in the network is independently hashed into a smaller set of variables.
    
    Reagen \etal~\cite{reagen2017weightless} use Bloomier filters \cite{chazelle2004bloomier} in order to index the weights. This work takes a post-training/pruning approach, the filters are not trained from scratch, and fine-tuning is needed to achieve good performance.  Similarly, Locality Sensitive Hashing has been used in \cite{Spring_2017} to maintain smaller weight pool. 
    
\vspace{-1em}
    \paragraph{Pruning}  \hspace{-1em} is the process of removing unnecessary weights~\cite{lecun1990optimal, liu2018rethinking} or entire neurons/filters~\cite{li2016pruning, tien2018netadapt, gordon2018morphnet} of the trained neural networks with the goal of maintaining as close as possible performance to the unpruned version. This can be achieved by penalizing the model with sparsifying norms~\cite{gordon2018morphnet, li2016pruning, CarreirIdelbay18a} or by ranking the weights/neurons~\cite{lecun1990optimal}, in one or multiple iterations.

 \vspace{-1em}
    \paragraph{Weight Quantization} Model quantization works by adjusting parameter values to lower precision \cite{Faraon_18a, Nagel2019DataFreeQT, wu2016quantized,Jacob2017QuantizationAT} or even binary weights \cite{courbariaux2015binaryconnect, Zhou_16b}. This has the desired effect of drastically reducing the size, and can be efficiently combined with pruning \cite{han2015deep, Choi_17a} to get even higher compression ratios.  \vspace{-1em}
    \paragraph{Decomposition} We can largely identify pure low-rank methods \cite{denil2013predicting, denton2014exploiting,Xu_18a} that apply matrix decomposition to fully connected and suitably reshaped convolutional weight matrices, and its generalization ---  tensor decomposition of layers \cite{Novikov_15a, Lebedev_15a, Garipov_16a, Wang_18b}.
    
\vspace{-1em}
    \paragraph{Architecture Design}
    A separate line of research is building compact models and training them from scratch \cite{Iandol_16a, mobilenetv1, tan2019efficientnet}, rather than compressing overparametrized ones.
    This is intrinsically manual process, and reinforcement learning methods are used to automate this task \cite{He_18a}.

\section{Method}
Our method is based on a hashing scheme applied to the original variables of the model.
It is inspired by Chen \etal~\cite{chen2015compressing} but rather than having a many-to-one mapping between weights and trainable variables, we use a many-to-many mapping. This exponentially reduces the probability of a full collision in the hash. Further, our approach maintains memory locality and so can be implemented efficiently, without latency overhead during inference.

Note that normally there is a 1-to-1 correspondence between the set of weight tensors of a model, and its set of trainable variables. In fact the names \emph{weights} and \emph{variables} are often used interchangeably. However, when considering model compression, specifically a hashing based approach, one needs to be clear about this distinction. Here we refer to \emph{weights} as the tensor values in convolution kernels, fully connected layers, biases and so on. We call \emph{variables} the set of trainable elements into which weights get hashed.

We denote by $\weight$ the weight tensor associated with layer $l$, and the elements of  $\weight$ as $[i_1,\ldots,i_k]$ for a rank $k$ tensor (e.g.\ $k=4$ for 2D convolutions, $k=2$ for fully connected layers).

For simplicity of notations, we will use $\weight[i]$ to cover both kernel-weights and biases.
We define $W=\{\weight\}^L_{l=1}$ the set of all weights of a network, and $|W|$ is the total number of weights of the model. These weights are essentially the set of all tensors which we set out to compress: convolution kernels, fully connected weight matrices, and biases, with the exception of the scale and bias parameters of batch-normalization\cite{ioffe2015batch}, as these parameters can be absorbed into the next operation during inference.

Here we should note that while it is common to hash or quantize weights \emph{after} training, we consider the use case of hashing weights into variables while the model is training. This allows the values of the hashing variables to be learned using back propagation.

\begin{figure*}[t]
\begin{subfigure}{.5\textwidth}
  \centering
  \includegraphics[width=.84\textwidth]{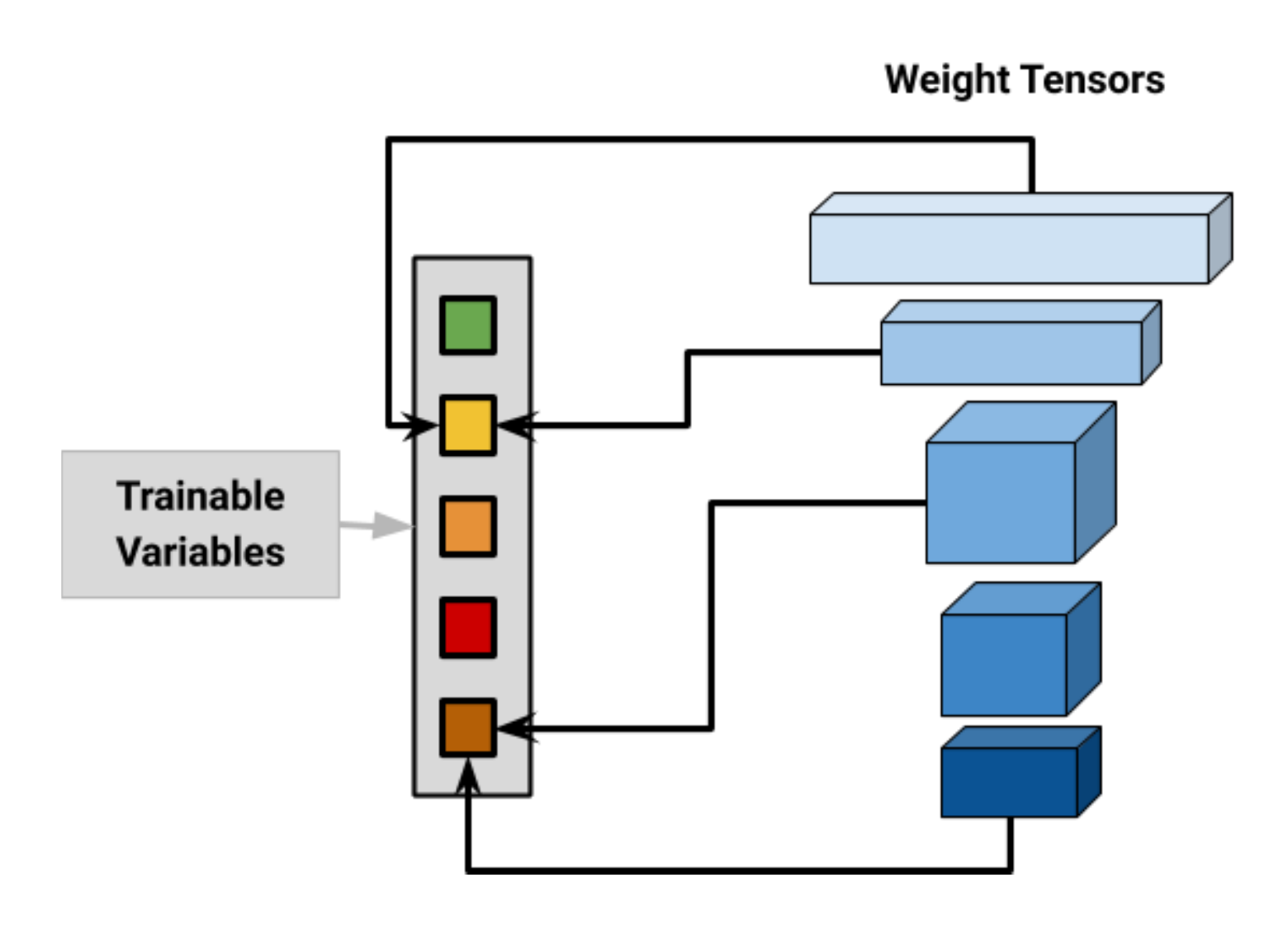}
  \caption{}
  \label{fig:simple-hash}
\end{subfigure}%
\begin{subfigure}{.5\textwidth}
  \centering
  \includegraphics[width=.78\textwidth]{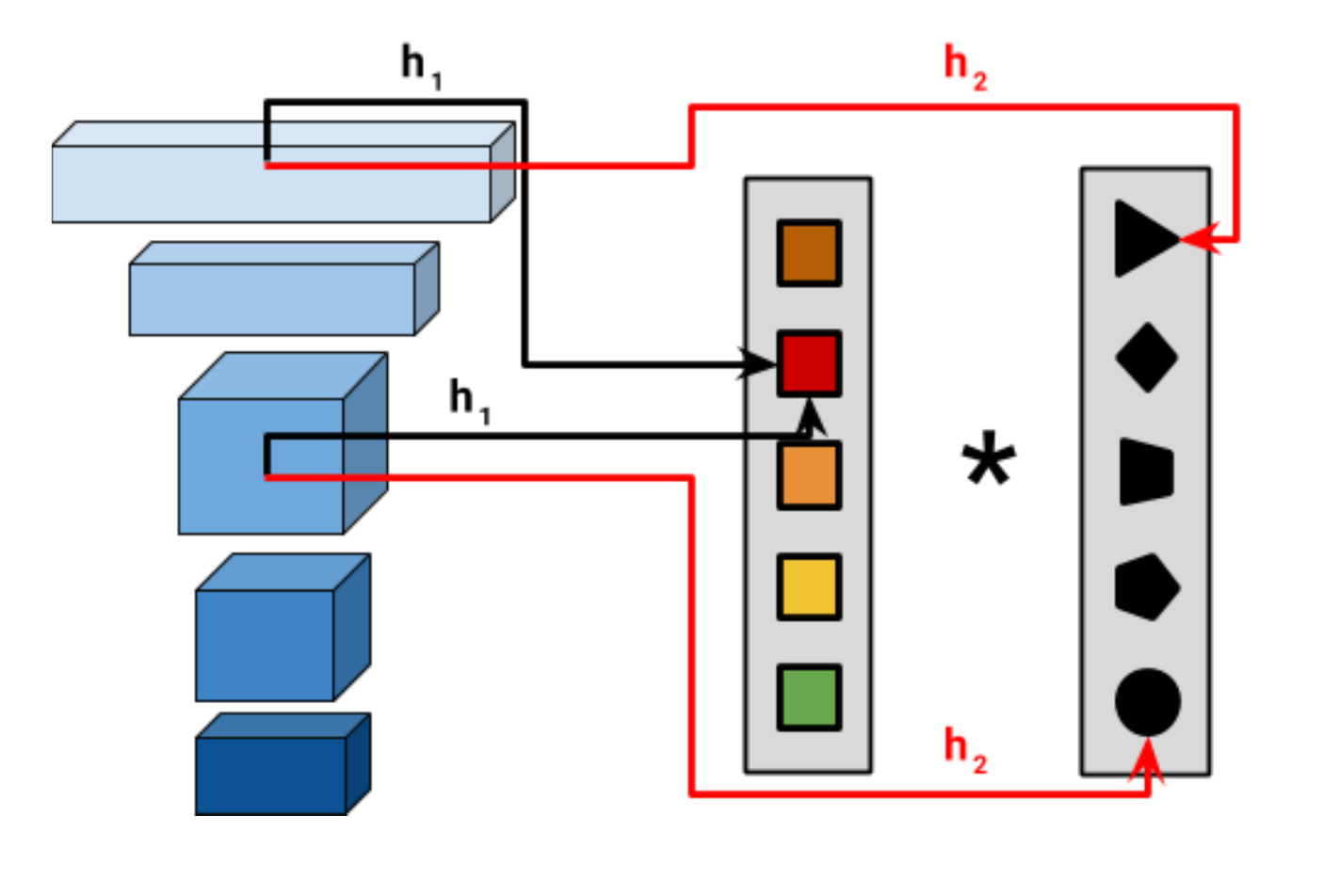}
  \caption{}
  \label{fig:multi-hash}
\end{subfigure}
\caption{(a) Model compression using hashing. Each value in the set of weight tensors in the model (blue cubes) is mapped into a single
    variable in the set of trainable variables. When strong compression is desired, many weight values are mapped into the same variable, creating equality constraints in the network. (b) Model compression using multi-hashing. Each value in the set of weight tensors in the model (blue cubes) is mapped into a number of trainable variables, each mapping using a different hash function. The probability of two weights being mapped to the same \emph{set} diminishes exponentially.}
\label{fig:fig-hashing}
\end{figure*}

\paragraph{Simple-hashing}
A simple hashing scheme is based on an underlying set of $\targetsize$ \emph{variables}, which we call a variable pool and denote by $\poolvar$.

The hashing is defined by a function $h$:
\[
h: [l,\vindex_l] \rightarrow \targetsize
\]
where $l$ takes values from $1$ to $L$ - the number of layers in the network, and $\vindex_l$ are indices into $\weight$.

The hash $h$ induces a mapping between model weights and variable: $\weight[\vindex] = \poolvar[h(l,\vindex)]$.
\ignore{Now any previous use in the network of $\weight$ will simply be using $\weight$s instead.}

The simple-hashing is similar to the one proposed in~\cite{chen2015compressing}, with the difference that it does not have random signs, and it operates on all the variables in the network (compared to their per layer approach). This simple-hashing scheme is shown in Figure~\ref{fig:simple-hash} and can be thought of training a network with a variable sharing pattern induced by the collisions of $h$. The number of collisions is equal to the number of weights reduced by the hashing scheme and when compression is not trivial, there is a large number of constraints. This could pose a problem if, for instance, a certain layer needs its weights to be of large magnitude, while another layer requires small values. When optimizing the model, an unfortunate compromise would arise. 

To overcome these hard collisions, we propose a multi-hashing scheme shown in Figure~\ref{fig:multi-hash} that induces a different set of constraints on the network in which the collisions induce softer, smoother, non-linear constraints.

\paragraph{Multi-hashing}
An $M$-hashing~\footnote{To clarify, we use the term multi-hash differently than commonly used in computer science theory - We employ our multi-hashes in \emph{parallel} to each weight index $\vindex_l$ to produce a set of variables, and then combine them using the reducer function and produce a value to be placed in $\weight[\vindex]$.}
is defined with a set of $M$ hash functions $\{h_m\}_1^M$ and $M$ variable pools $\{\poolvar_m\}_1^M$, and a reducer function $\phi : \reals^M \rightarrow \reals$.

These define the following mapping between model weights and variables: 
\[
    \weight[\vindex] = \phi(\poolvar_1[h_1(l,\vindex)], \ldots, \poolvar_M[h_M(l,\vindex)]) 
\]

The choice of the reduction function is an important component in the multi-hashing scheme. The sum function is an example of a simple reducing function: 
\begin{equation}\label{eq:sum-reducer}
\phi(x_1,...x_M) = \sum_{m=1}^M x_m
\end{equation}
other functions such as the product, min, or max can also be considered.

\subsection{Structured Multi-Hashing}
Instead, we use multi-hashing to partition the variables into groups which share some dependency structure. Common hash functions would create random partitioning, but this loses a property which could be important for our use case: memory locality. Neighboring weights in the network can be mapped to arbitrary variables in the pool and so a layer $l$ potentially needs to access all the variable pools to compute its output.

Specifically, when we consider an implementation where we do not unpack the hashing offline, but compute the values of the weight tensors on the fly, then the cost of fetching all the variable pools could be significant. We propose the notion of structured hashing which will increase the memory locality.
\ignore{(we can explain that in batch size 1, and compression ratio $>10\%$ we will see a high memory payload for later layers)}

We define $2M$ sum-product reducer as:
\begin{equation}
    \phi_{\sum\prod}(x_1,...x_{2M}) = \sum_{m=1}^{M} x_{2m-1}x_{2m}
\end{equation}
Combined with a carefully chosen hashing scheme, this reducer will maintain memory locality and is efficient during inference.

\begin{figure}[t]
  \centering
    \includegraphics[width=0.47\textwidth]{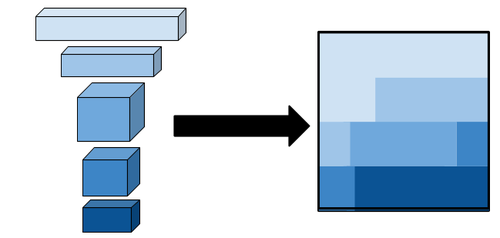}
    \caption{We conceptually represent the full set of weights of a deep model as a single, square, matrix. In this matrix $\weight[\vindex]$ is mapped to some coordinates $M_W[r, k]$. We then apply structured multi-hash to represent this matrix using a sum-product of smaller matrices.}
    \label{fig:representation}
\end{figure}

First we re-parameterize the way we refer to the weights.
Let $N=|W|$ be the number of total weights in the network. Rather than $\weight[\vindex]$ we think of all the weights of a model as if coming from a single square matrix of dimension $n=\lceil \sqrt{N}\rceil$. The mapping between the weights and the elements of the matrix is trivially achieved by tiling the weights in the order of their creation: 
\begin{equation}\label{matrix-mapping}
    \weight[\vindex] = \matvar[r_\vindex, c_\vindex]
\end{equation}
where $r_\vindex$ and $c_\vindex$ are row and column indices determined by $\vindex$. This mapping is illustrated in Figure~\ref{fig:representation}.

The core idea of our structured hashing approach is to encode the locality of weights using matrix operations.
A multi-hashing scheme determines how we \emph{represent} $M_W$. We define a $2M$ Matrix-Product Multi-Hashing:
\begin{equation}\label{matrix-product-hash}
    \matvar = [\poolvar_1;\poolvar_2; \ldots \poolvar_{M}]^t [\poolvar_{M+1};\poolvar_{M+2};\ldots \poolvar_{2M}]
\end{equation}
Where $\poolvar_i$ are column vectors of size $n$.
Note that the re-parameterization defined in Eq.~\eqref{matrix-mapping} combined with the decomposition into hash functions in Eq.~\eqref{matrix-product-hash} implements a $2M$ multi-hash with a sum-product reducer in a way that is memory efficient.

Note that our approach computes a low rank approximation of the weight matrix of the entire model. However, this does not assume that any specific layer is low rank nor does it enforce it.

\subsection{Selecting the Number of Hashes}
The size of the Matrix-Product Multi-Hash, i.e.\ the number of trainable variables it creates is $2Mn$. This is determined by the size of the hash vectors $\poolvar$ and the number of hashes. These define the size of the matrices in Eq.~\eqref{matrix-product-hash} -- $n \times M$ and $M \times n$. We can hash the model into any target size $T$ (up to rounding errors in the order of $\sqrt{N}$) by setting~$M= \left \lceil {T}/{2n} \right \rceil$.

\subsection{Scaling and Initialization}
Correctly initializing the weights of a deep network is often important for it to train well.
As such this is an active area of research and there is a plethora of initialization methods available to practitioners, and each model architecture is paired with an initialization scheme that fits it. For our multi-hash compression method, to match the performance of the uncompressed model, we would want the weights to be initialized using a matching distribution. Note however that a single weight value in our method is the sum-of-products of $2M$ variables. For a target distribution $\mathbb{D}$ one can define $2M$ distributions $\mathbb{d}_m$ such that the distribution of their sum-product is equals $\mathbb{D}$. Specifically, for the commonly used Gaussian distribution the sum part is trivial as the Gaussian family is closed to additions. However, although well defined, a distribution where its product is a Gaussian is hard to sample from \cite{pinelis2018exp}. 
Instead, we focus on matching two properties of $\mathbb{D}$: its range, and scale. Note that the common practice in deep models is to use the same family of distributions in all layers, but with an appropriately selected per-layer-scale. We follow that practice here.

\vspace{-1em}
\paragraph{Range} Initialization schemes can be categorized into two types: unbounded distributions (e.g.\ Normal), and bounded ones (e.g.\ uniform, truncated normal). When initializing variables in the pool we match the range property.
 
\vspace{-1em}
\paragraph{Scale} The challenge with the scale is that different layers can be initialized to different scales. This happens for instance with Glorot \cite{glorot2010understanding} initialization where the standard deviation is a function of the fan-in and fan-out of the layer.
In this setting, it is impossible to initialize the hash variables such that all layers simultaneously have the desired scale. We solve this by first initializing variables so $std(\weight) = 1$, and then we re-scale each layer to match the target scale ${s}^l$. For the sum-product reducer we standardize the resulting weights by setting the standard deviation $\sigma$ or the underlying variables according to:
\begin{align*}
    std(\weight) & = std \left ( \sum_{m=1}^{M} x_{2m-1}x_{2m} \right ) \\
    & = \sqrt{\sum_{m=1}^{M}  var(x_{2m-1})var(x_{2m})} \\
    &= \sqrt{M\sigma^4}
\end{align*}
Setting $\sigma =M^{-\frac14}$ creates weights with unit standard deviation. Then, multiplying by ${s}^l$ allows us to effectively control the scale of each layer.

\subsection{Per-Layer Learnable Scale}\label{sec:per-layer-scale}
While our multi-hash technique removes equality constraints, there are still dependencies between weights as they share some of the variables used in their sum-products. Consider two layers $l$ and $l'$, it could be hard for the network to learn different scales for $w^{l}$ and $w^{l'}$ due to sharing of the underlying variables. The per-layer scaling mentioned above addresses this problem at initialization, but layers initialized with the same scale are bound to keep similar scales while training. We allow the per-layer scale to be a \emph{learnable} variable, which provides the network with another degree of freedom to address this issue. Our experiments in Sec~\ref{sec:res-per-layer-scale} show that this small set of extra variables (one per layer) are always helpful, and result in $0.5\%-1\%$ improvement in accuracy.

\section{Results}
In this section, we evaluate our method on three model families: ResNets, EfficientNets, and MobileNets. 
We show that \SMASH\ compression can drastically reduce model size with minimal loss in quality. In fact, when compressing large models, we often outperform comparably sized models from the same family.

\subsection{ResNet Models}
ResNet architectures~\cite{he2016deep} are versatile and so are used in many applications. They are also popular as benchmark models. There are two main procedures used to make ResNet models cheaper:
Changing the number of \emph{layers} (e.g. ResNet101, ResNet50, ResNet18) and changing the number of \emph{filters}, usually done with uniform scaling and is commonly known as width multiplier.

\begin{figure}[t]
  \centering
    \includegraphics[width=0.47\textwidth]{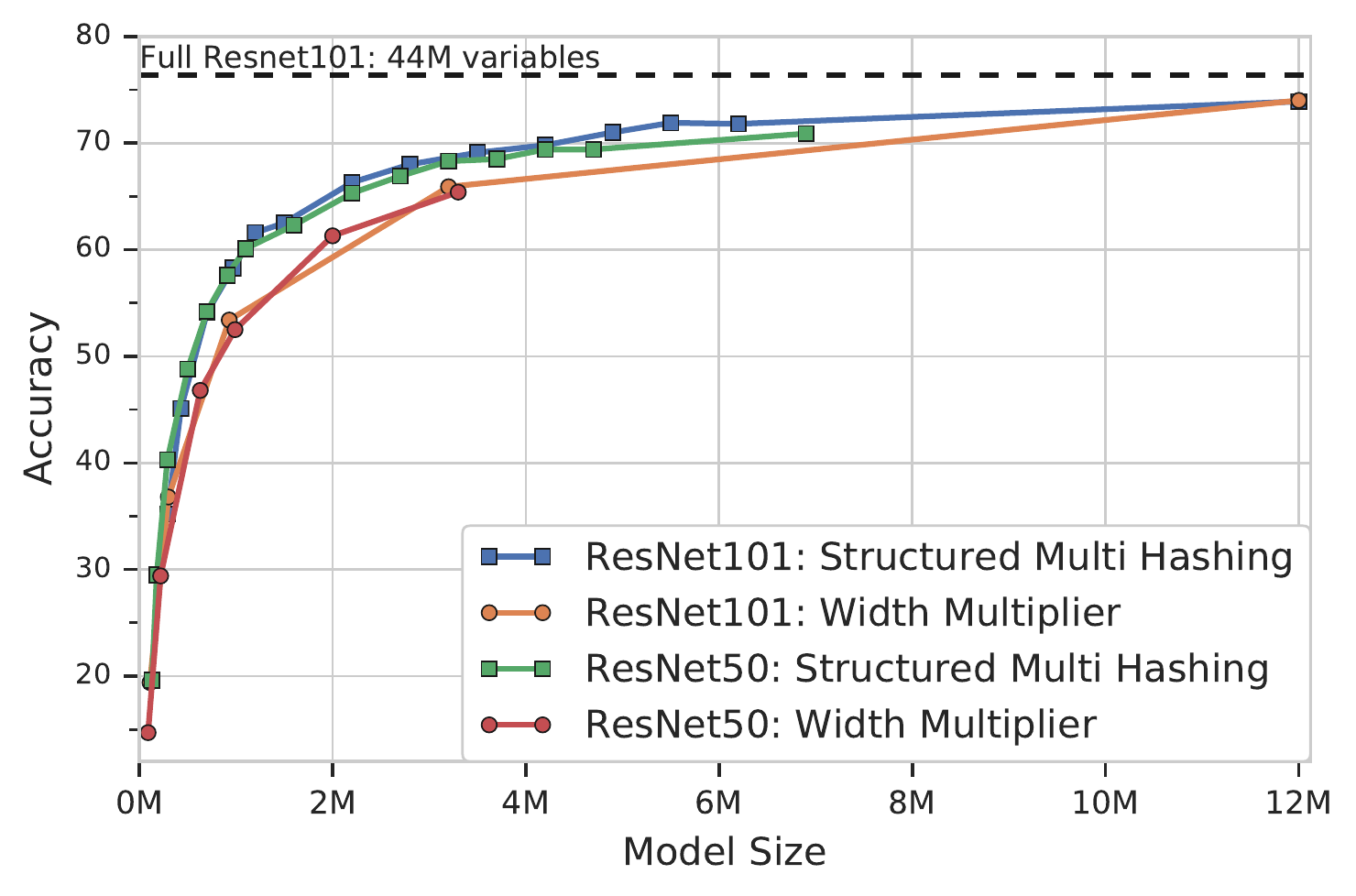}
    \caption{Accuracy vs Model Size with ResNet based models on Imagenet.
    We compare \SMASH\ to shrinking using Width Multiplier on ResNet50 and ResNet101.
    For both model types, \SMASH\ finds better tradeoffs between size and accuracy.
    }
    \label{fig:resnet-accuracy}
\end{figure}

Figure~\ref{fig:resnet-accuracy} shows our structured multi-hashing compression on ResNet50 and ResNet101. Each point on the curve is one
model trained to convergence to a specific target size. We compare with shrinking each one of the models using width multiplier. Note that SMH compresses the model more efficiently. For example, for an accuracy of 70\% SMH models are half the size of the width multiplier models.

\subsection{EfficientNet Models}
The family of EfficientNet models~\cite{tan2019efficientnet} provides a natural and strong baseline for comparison. Using a large scale study of model hyper-parameters that affect size and latency, they propose a model scaling formula. By applying this formula, the authors 
propose 8 different models spanning from very large (B7) to very small (B0).

\begin{figure}[t]
  \centering
    \includegraphics[width=0.47\textwidth]{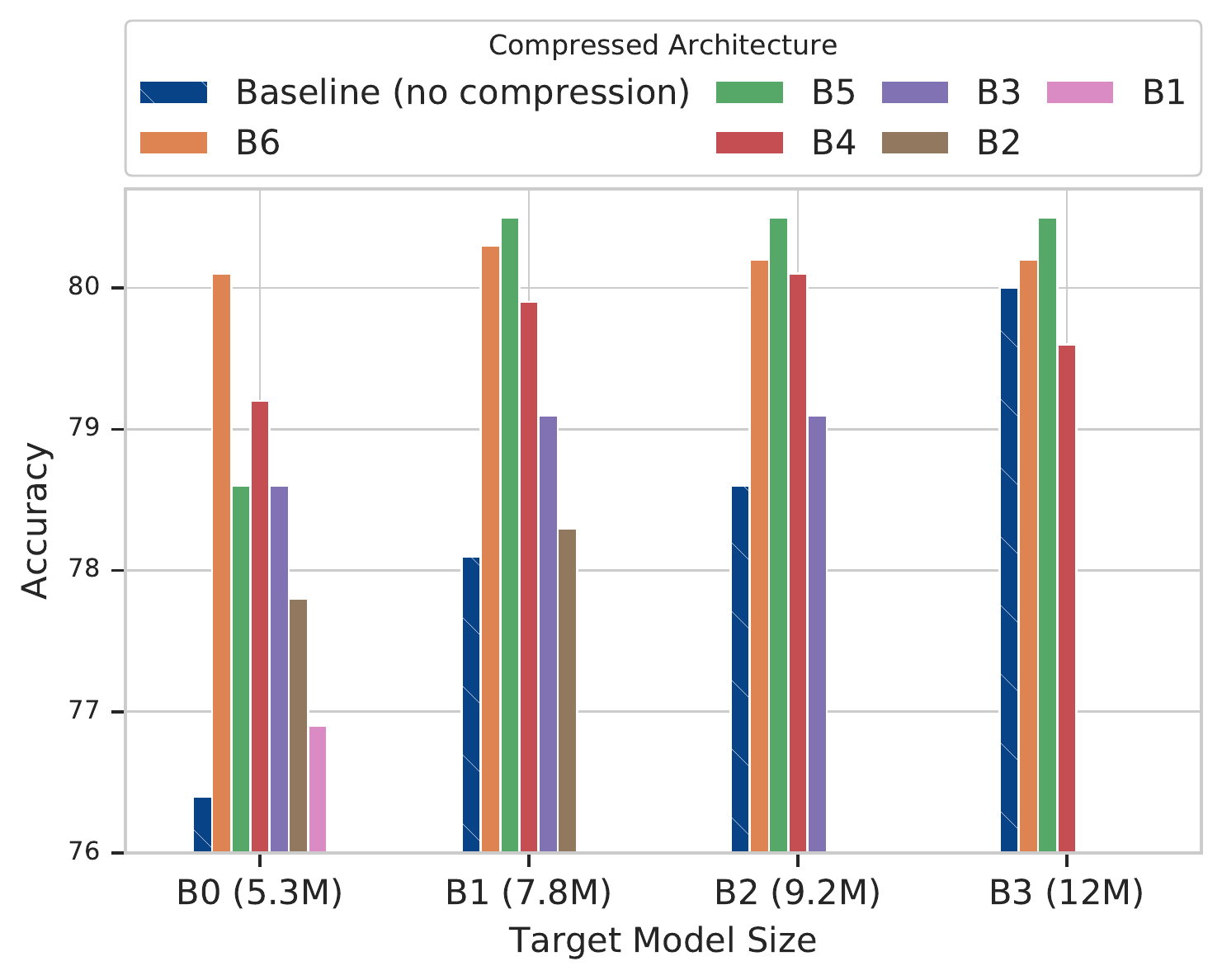}
    \caption{Accuracy vs Target Model Size on EfficientNet. Bars represent original architecture used. They are grouped by the size of the model after compression. For example, compressing a B6 architecture to the size of B0 has an accuracy of $80.1\%$ compared with the original B0 at $76.4\%$. Note that the B5 and B6 architectures are slow to train and are harder to find stable hyper-parameters.}
    \label{fig:target-model-size}\vspace{-0.2cm}
\end{figure}

To evaluate the merit of our compression technique, we apply it to a subset of the EfficientNet family (namely B0 to B6). For each model, we use the size of smaller variants as target sizes. For example, we compress the B3 architecture to 5.3M, 7.8M, and 9.2M parameters corresponding to the sizes of B0, B1 and B2 variants. 
Figure~\ref{fig:target-model-size} shows the results of applying this procedure. The bars are grouped by the target size to which they are compressed. Each bar indicates a starting model architecture.

Note that we can significantly improve accuracy for any desirable model size compared with the original model. For example, the original B0 model has an accuracy of $76.3\%$, but a B6 architecture compressed to the size of B0 has $80.3\%$.
Even more drastic is comparing between groups --- a B4 model compressed to the size of B1 outperforms the original B3 model even though it is $35\%$ smaller.

\subsection{MobileNets}
MobileNets \cite{mobilenetv1, mobilenetv2, mobilenetv3} are a family of models specifically targeted to mobile devices. These models have been primarily optimized  for FLOPs. However they are also significantly smaller than other models considered above.

\begin{figure*}[t]
\begin{subfigure}{.47\textwidth}
  \centering
  \includegraphics[width=1.0\textwidth]{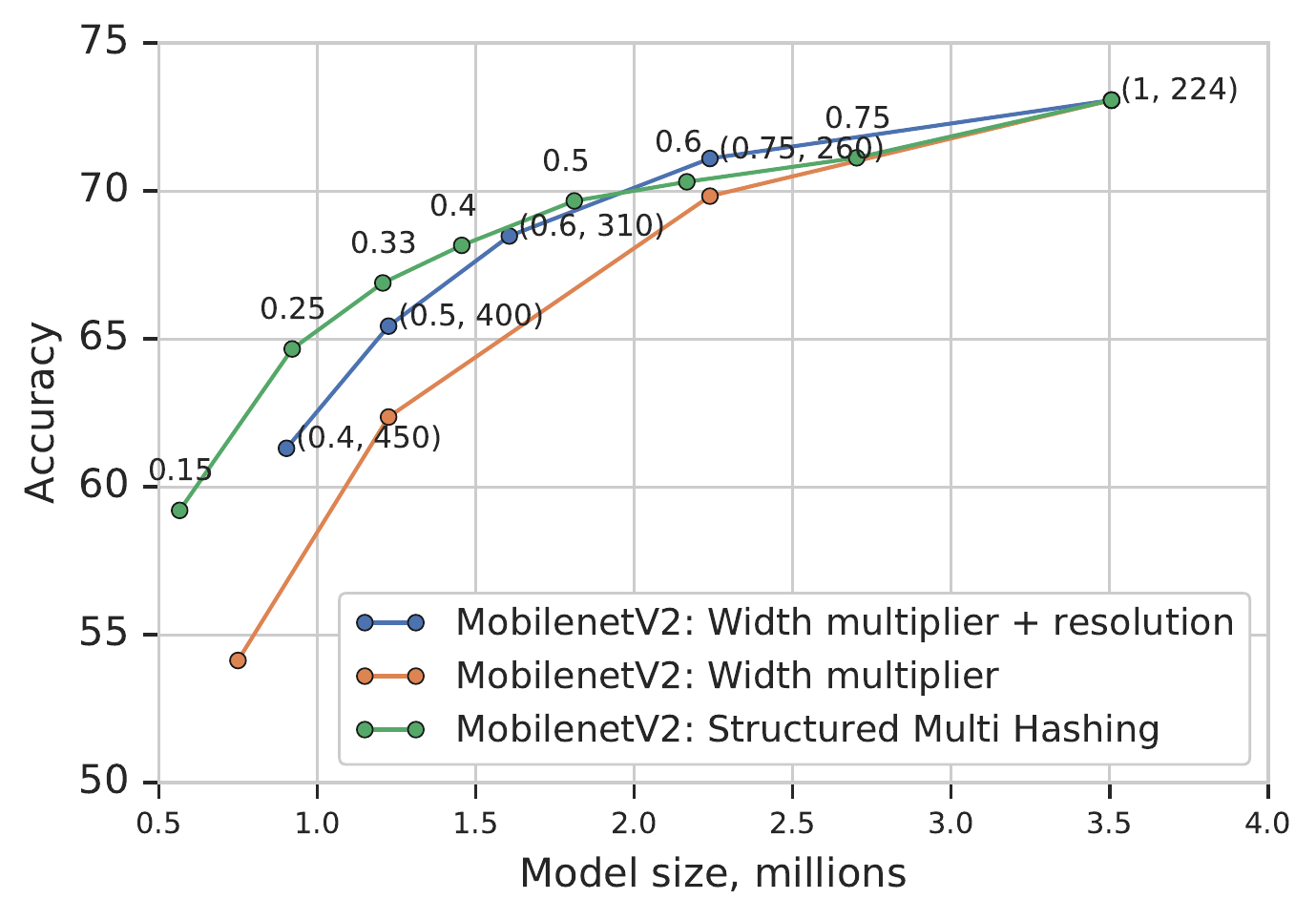}
  \caption{}
  \label{fig:mobilenet-v2}
\end{subfigure}%
\begin{subfigure}{.47\textwidth}
  \centering
  \includegraphics[width=1.0\textwidth]{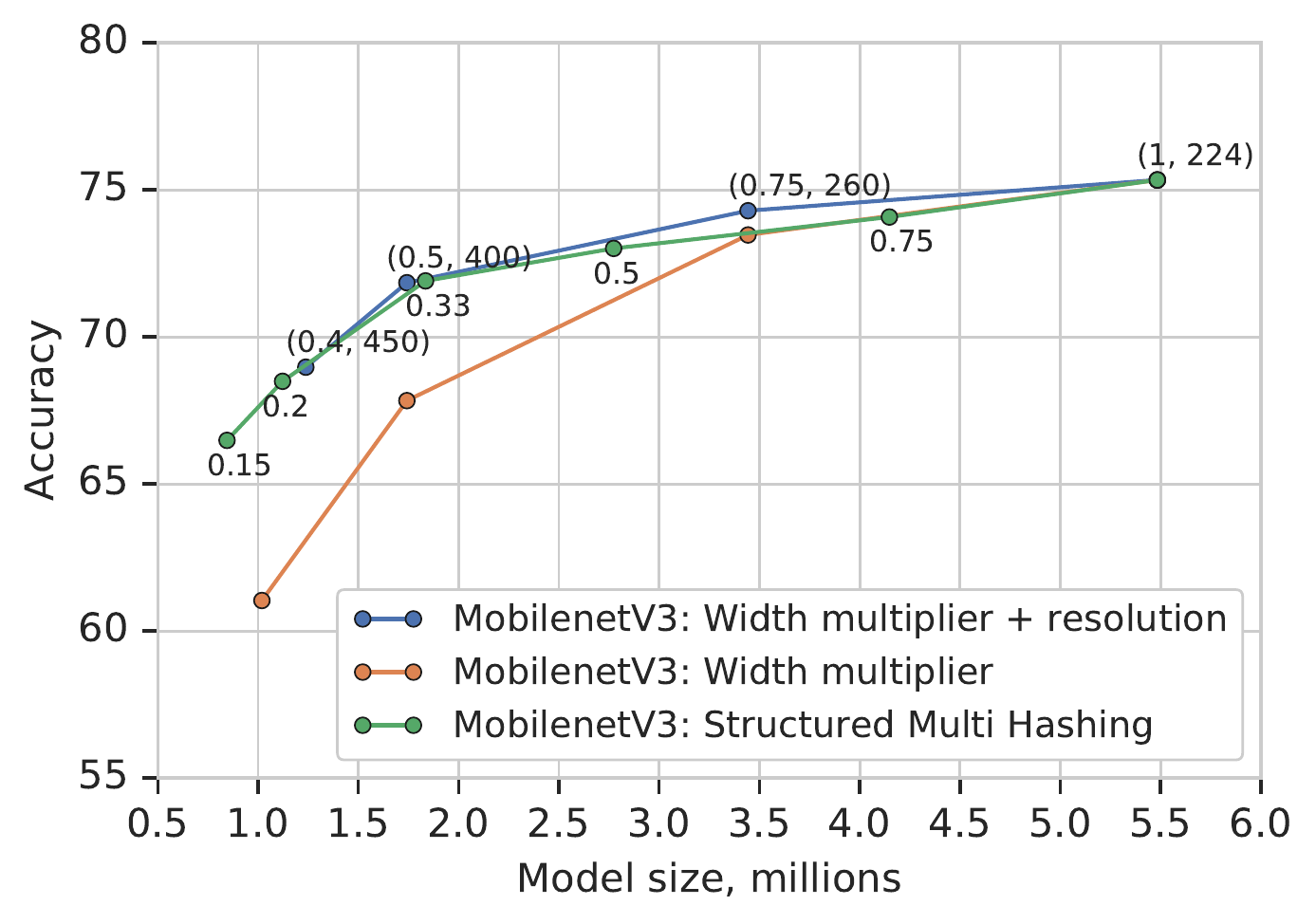}
  \caption{}
  \label{fig:mobilenet-v2}
\end{subfigure}
\caption{Applying \SMASH\ to  MobilenetV2 (a) and MobilenetV3 (b) on ImageNet. 
    We compare to two baselines. Width-multiplier: all layers are shrunk by a constant factor. Width-multiplier + resolution is a stronger baseline where we additionally up-sample the input image to maintain the same
    number of FLOPS. Label for strong baseline such as  (0.4, 450) represent multiplier and up-sampled resolution respectively. Label for \SMASH~ is the target compression rate.}
\label{fig:mobilenet}
\end{figure*}

In this section, we measure the efficacy of applying structured multi-hashing to MobileNetV2 and MobileNetV3. The comparison is presented on Figure~\ref{fig:mobilenet}.
In addition to the width-multiplier as we did for ResNet, we also impose an additional, stronger baseline based
on a combination of width-multiplier and resolution multiplier.

Width-multiplication reduces both FLOPs and model size, while structured-hashing only reduces the model size. To make a stronger baseline that produces comparable FLOPs, when we apply width multiplier $\alpha$ to reduce the model size, we increase resolution by $\approx \alpha$ that brings FLOP count back to the original cost. 

Note, in contrast with~\cite{tan2019efficientnet}, and following~\cite{weakmodel}, we don't actually use higher resolution image. Instead, we simply up-sample the input. This guarantees that all models are trained on exactly the same data. It is interesting that for MobileNetV2, the multi-hash approach beats both baselines. On the other hand, for MobileNetV3, the stronger baseline produces slightly better trade-off curve around the full model. However, we note that the strong baseline is both slower and requires more memory to train (due to high spatial resolution of early tensors). In fact, we were unable to train the strong baseline example with multiplier less than 0.4, which required using input upsampled to 450x450 . Another potential issue that limits the usefulness of the strong baseline is that it requires fractional upsampling which introduces image artifacts.

\subsection{Compared to Non Structured Hashing}
Here we compare the results of \SMASH\ to a single and multi-hashing baselines. We implement model hashing over the full set of network weights. For standard hashing we hash each weight into $K\in \{1, 2, 10\}$ sets of trainable variables and use the sum reducer defined in Eq.~\eqref{eq:sum-reducer}. We compare the methods on an EfficientNet B2 model compressed to 5M and 7.9M trainable variables (the size of B0, and B1 respectively). Table~\ref{tab:compare-hash} shows the results. \SMASH\ is both more accurate and much faster then standard hashing.
The memory locality of \SMASH\ and its implementation as a matrix product result in this low overhead. Note that \SMASH\ in these experiments is using $\sim 800$ hash functions.

\begin{table}
\small
\begin{tabular}{cccc}
\toprule
Compression & Target Model &  Accuracy &  Samples\\
Method &  Size &   &  Per Second \\
\midrule
            \SMASH &   5.3M &     \textbf{0.774} &       6060 \\
           1X Hash &   5.3M &     0.762 &                4000 \\
           2X Hash &   5.3M &     0.765 &                2800 \\
          10X Hash &   5.3M &     0.770 &                 790 \\
\midrule
            \SMASH &    7.9M &    \textbf{0.782} &        6040 \\
           1X Hash &    7.9M &     0.773 &                3900 \\
           2X Hash &    7.9M &     0.775 &                2500 \\
          10X Hash &    7.9M &     0.779 &                 760 \\
\bottomrule
\end{tabular}
\caption[]{Hashing methods comparison on EfficientNet B2 model. We compare \SMASH\ to non-structured hashing with 1,2  and 10 hash functions. Note that not only is \SMASH\ more accurate but also much faster. We measure training samples per second on TPU V3 with a 4x4 topology. 
}
\label{tab:compare-hash}
\end{table}

\subsection{Extreme Compression}
As noted above, deep networks are known to be over-parameterized. Here, we examine this notion further. We ask the question: \emph{Can deep models be accurate when using an extremely small number of trainable variables?} Can this be done for an architecture that was not specifically designed for this purpose? To answer this question we perform two sets of experiments, on CIFAR10~\cite{krizhevsky2009learning} using a ResNet32 model, and on ImageNet using EfficentNet models. 

\begin{figure}[t]
  \centering
    \includegraphics[width=0.47\textwidth]{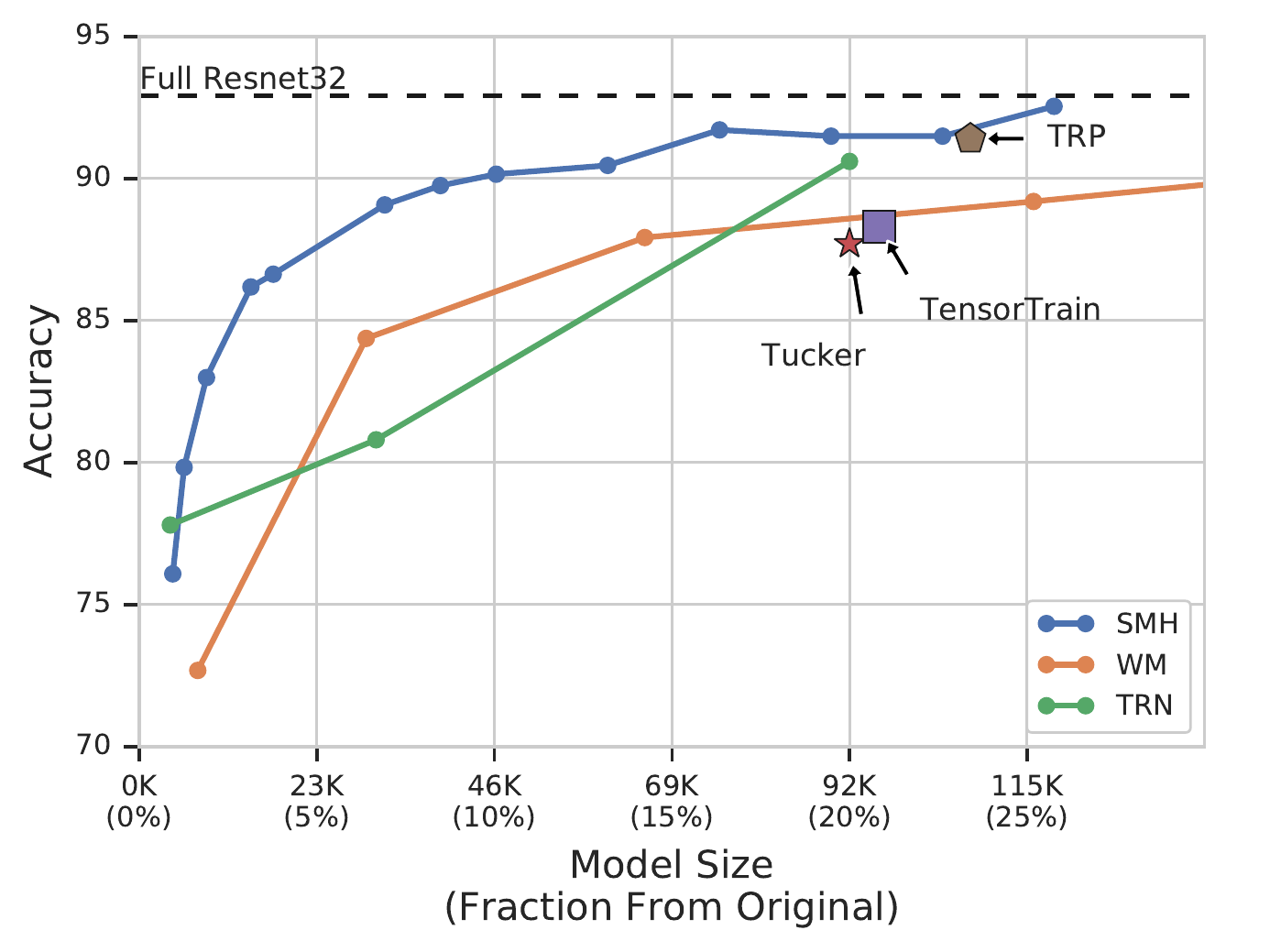}
    \caption{Accuracy vs Model Size on CIFAR10. We compare a number of compression methods on the ResNet32 model. We focus on the extreme compression regime. The rightmost points represent models that are $25\%$ of the original ResNet32. Here TRP is low-rank ResNet32 from \cite{Xu_18a}, Tucker, Tensor-Train and Tensor-Ring-Net (TRN) results obtained from \cite{Wang_18b}, WM is width multiplier.}
    \label{fig:resnet32-cifar10}
\end{figure}

In Figure~\ref{fig:resnet32-cifar10} different compression method applied to a ResNet32 model trained on the CIFAR10 dataset are show. First note that using our multi-hash approach we can effectively discard $75\%$ of the variables in the model, without loss in performance. Furthermore, we can create a model with only $10\%$ of the original size (only $50K$ variables) and still maintain an accuracy above $90\%$. 

Secondly, we take EfficentNet B4 and B5 architectures and compress them using \SMASH\ by 10x to 2M and 3M parameters respectively. In Table~\ref{tab:extream-eff} we compare them to vanilla EfficentNet models with similar accuracy and see accuracy gains in both cases. 

\begin{table}
\small
  \centering
\begin{tabular}{lcc}
\toprule
Model & Accuracy & Model Size \\
\midrule
B0 & 76.3\% &5M \\
\SMASH$_{2M}$ B4  & 76.6\% & 2M \\
\SMASH$_{3M}$ B5  & 78.3\% & 3M \\
B1 & 78.8\% &7.9M \\
\bottomrule
\end{tabular}
\caption[]{\SMASH\ B4 model is 60\% smaller than B0 with the same accuracy. \SMASH\ B5 model is 40\% smaller and 2\% better than B0 and 63\% smaller than B1 with slightly lower accuracy.}
\label{tab:extream-eff}
\end{table}

\subsection{Per-Layer Learnable Scale}\label{sec:res-per-layer-scale}
To examine the benefits of adding a per-layer scale variable, as described in Sec~\ref{sec:per-layer-scale}, we train 16 EfficientNet based models on ImageNet. We train five different base models B0 to B4, and for each base model set a number of target sizes to compress to. We then train the models until convergence with and without the per-layer scale variable using the same hyper-parameters.

Table~\ref{tab:per-layer-scale} shows the accuracy difference in accuracy when adding per-layer scale variables. We usually see improvements of about $0.5\%$ to $1\%$. Also note that this procedure never hurts performance.
\begin{table}[t]
\small
\centering
\begin{tabular}{ccccc}
\toprule
Base  & Target & Fixed       & Learnable & Accuracy \\
Model & Size   & Scale    & Scale     & Difference \\
\midrule
    B0 & 2.0M &    71.9\% &           \textbf{73.1\%} &       1.2\% \\
       & 3.0M &    73.7\% &           \textbf{74.0\%} &       0.3\% \\
\midrule
    B1 & 2.0M &    73.4\% &           \textbf{74.2\%} &       0.8\% \\
       & 3.0M &    75.2\% &           \textbf{76.5\%} &       1.3\% \\
       & 5.0M &    76.4\% &           \textbf{76.8\%} &       0.4\% \\
\midrule
    B2 & 2.0M &    73.9\% &           \textbf{74.9\%} &       1.0\% \\
       & 3.0M &    75.7\% &           \textbf{77.1\%} &       1.4\% \\
       & 5.0M &    77.0\% &           \textbf{77.8\%} &       0.8\% \\
\midrule
    B3 & 2.0M &    75.1\% &           \textbf{75.5\%} &       0.4\% \\
       & 5.0M &    78.0\% &           \textbf{78.6\%} &       0.6\% \\
       & 7.9M &    78.4\% &           \textbf{79.1\%} &       0.7\% \\
       & 9.3M &    78.2\% &           \textbf{79.1\%} &       0.9\% \\
\midrule
    B4 & 5.0M &    78.8\% &           \textbf{79.2\%} &       0.4\% \\
       & 7.9M &    79.2\% &           \textbf{79.7\%} &       0.5\% \\
       & 9.3M &    79.3\% &           \textbf{79.5\%} &       0.2\% \\
\bottomrule
\end{tabular}
\caption{Accuracy improvement when adding a per-layer learnable scale. Average accuracy gain is $0.7\%$, note that the per-layer scale is \emph{always} beneficial.}
\label{tab:per-layer-scale}
\end{table}

\subsection{Targeted Weight Compression}
When compressing a neural network, one can choose to target all weights, or a smaller subset of them. For example, in all our experiments we do not to hash any of the Batch Normalization variables, as those can be absorbed in the following convolution during inference.

\begin{figure}[t]
  \centering
    \includegraphics[width=0.47\textwidth]{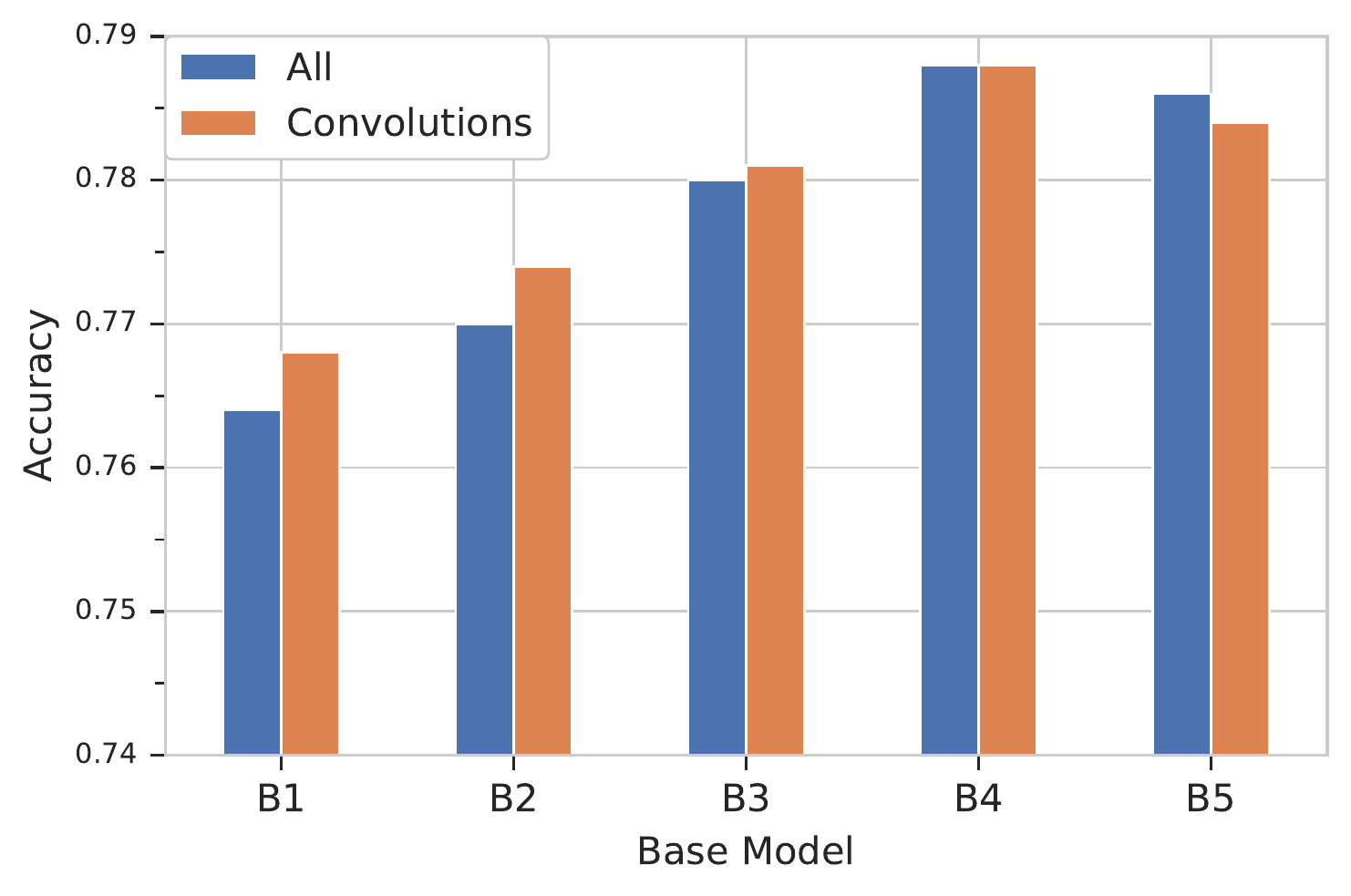}
    \caption{Compressing EfficientNet architectures to the size of B0 (5M).
    Bars are grouped by the architecture being compressed. We compare compressing all layers (blue), with hashing only convolutional layers (orange). There is a small difference - the method isn't sensitive to this design choice. Note that smaller models benefit from maintaining the fully connected layers untouched, while the larger base models gain from the flexibility to compress all layers.}
    \label{fig:sensitivity}
\end{figure}

One natural distinction between model weights is to separate those coming from convolutional layers which are usually in the early stages of a model, and those coming from fully connected layers which are commonly at the later part of the network. Figure~\ref{fig:sensitivity} shows five base architectures from the EfficientNet family (B1, ..., B5) all compressed to the size of B0 (5M variables).
For each base model we once compress it by hashing all the weights, and once by only hashing weights coming from the convolutions.

The differences are not big, indicating that the multi hashing constraint has enough flexibility to make useful trade-offs. Note also that when starting with smaller architectures (B1, B2) it is better to limit the hashing to the convolutions. The fully connected layers in those models are smaller and have less representation power to spare.

When starting with bigger models however (B4, B5) the trends reverses and higher accuracy is achieved when letting the multi-hash compress all layers. For these models, the fully connected layers have many parameters and without access to those layers the hash must compress the rest of the network more drastically. For example, the B5 architecture has 2M weights in its dense layer, out of 30M. When compressing it to the size of B0 without hashing the dense layer we now need to hash 28M weights into 3M variables, a $89\%$ compression of those layers.

\section{Discussion}
In this paper we have presented an efficient model compression method that builds on the idea of weight hashing, while addressing its key limitations:
We eliminate hash collisions by introducing a multi-hash and reduce framework which maps each weight in a model into a set of trainable variables, and computes its value using a reduce operation on the set. Memory locality is preserved by eschewing random hashing, and defining a structured mapping instead. The \SMASH\ approach can be represented as a matrix product and does not add material overhead to model latency.

We show that a well optimized hashed model can be strongly compressed with minimal loss in accuracy. We demonstrated our results on the widely used ResNet family of models, and on the newer and more powerful EfficientNet and MobileNet model family.

From a scientific perspective, model hashing is distinctly different from quantization or pruning. Model quantization changes the precision in which the underlying function is approximated, but does not change 
dimensionality of the approximator. Pruning induces some weight values to zero but this on its own has no effect on the overall dimension.
If the pruning is strong enough to set complete rows of weight matrices to zero, or if it has a structured form, e.g.~\cite{gordon2018morphnet} it changes both the dimensionality of the approximator (number of variables) and its expressivity (number of layers, amount of non-linearity, etc).
In contrast model hashing does not change precision, but affects only the dimension (i.e. number of variables). 

Model hashing then provides a useful tool for exploring the role of the number of variables within an architecture family. Our results on the
ResNet family of models shows that number of variables tracks closely
with accuracy. ResNet101 and ResNet50 based models, compressed to the
same number of parameters perform almost indistinguishably from each other. This is true both for our hashing technique, and for the width multiplier baseline.

In contrast this does not hold for the EfficientNet or Mobilenet
family, in which different architectures (e.g.\ B6 vs B4, or V2 vs V3) compressed to the same size, differ significantly in their accuracy. Clearly more work is needed before we can fully understand the role of parameter counts.

\end{document}